\documentclass[aps,pre,twocolumn,showpacs,preprintnumbers,amsmath,amssymb]{revtex4}
\usepackage{graphicx}
\usepackage{epstopdf,xcolor,soul,multirow}
\usepackage{dcolumn}
\usepackage{bm}
\usepackage{mathrsfs} 
\usepackage{amsmath} 
\usepackage{xr-hyper}
\usepackage[colorlinks=true,linkcolor=blue,citecolor=blue]{hyperref}%
\usepackage{graphicx}
\usepackage[utf8]{inputenc}
\usepackage[ruled,vlined]{algorithm2e}
\usepackage{hyperref}
\usepackage[utf8]{inputenc}
\usepackage[T1]{fontenc}
\usepackage{mathptmx}
\usepackage{xcolor}

\newtheorem{remark}{Remark}
\begin{document}	
\title{Optimized ensemble deep learning framework for scalable forecasting of dynamics containing extreme events}
\author{Arnob Ray$^1$}\thanks{Equal contribution}\email{arnobray93@gmail.com}
\author{Tanujit Chakraborty $^2$}\thanks{Equal contribution}\email{tanujitisi@gmail.com}
\author{Dibakar Ghosh$^1$}\email{dibakar@isical.ac.in}
\affiliation{$^1$Physics and Applied Mathematics Unit, Indian Statistical Institute, Kolkata 700108, India\\
	$^2$Center for Data Sciences, International Institute of Information Technology Bangalore 560100, India}
\date{\today}

\begin{abstract}
The remarkable flexibility and adaptability of both deep learning models and ensemble methods have led to the proliferation for their application in understanding many physical phenomena. Traditionally, these two techniques have largely been treated as independent
methodologies in practical applications. This study develops an optimized ensemble deep learning (OEDL) framework wherein these two machine learning techniques are jointly used to achieve synergistic improvements in model accuracy, stability, scalability, and reproducibility prompting a new wave of applications in the forecasting of dynamics. Unpredictability is considered as one of the key features of chaotic dynamics, so forecasting such dynamics of nonlinear systems is a relevant issue in the scientific community. It becomes more challenging when the prediction of extreme events is the focus issue for us. In this circumstance, the proposed OEDL model based on a best convex combination of feed-forward neural networks, reservoir computing, and long short-term memory can play a key role in advancing predictions of dynamics consisting of extreme events. The combined framework can generate the best out-of-sample performance than the individual deep learners and standard ensemble framework for both numerically simulated and real world data sets. We exhibit the outstanding performance of the OEDL framework for forecasting extreme events generated from Li{\'e}nard-type system, prediction of COVID-19 cases in Brazil, dengue cases in San Juan, and sea surface temperature in Ni{\~ n}o 3.4 region.
\end{abstract}

\maketitle

\section{\label{sec:intro} Introduction}
The study of extreme events is one of the interdisciplinary research fields due to its devastating impact on nature and human civilization.
 Several disciplines deal with this topic of extreme events due to their occurrence in various fields \cite{hobsbawm1995age,albeverio2006extreme}. We can highlight a few names of extreme events such as the spreading of pandemic COVID-19 in the whole world \cite{machado2020rare}, 2020 global stock market crash \cite{mazur2021COVID}, super cyclonic storm Amphan in India \cite{hassan2020quantitative}, oceanic rogue wave near Newfoundland \cite{web_2}. Last few decades, researchers are interested to explore extreme events from dynamical system approach \cite{lucarini2016extremes}. 
 Extreme events occur in a dynamical system when its trajectory evolves within a bounded region most of the time, but occasionally moves far away from that region and originates a significantly large amplitude for a while in temporal dynamics \cite{sapsis2018new}. 
 Several experimental and numerical works have been done based on the appearance and origination of extreme events in various dynamical systems \cite{bonatto2011deterministic, pisarchik2011rogue, zamora2013rogue, kingston2017extreme, ansmann2013extreme}. However, unpredictability is an important attribute of extreme events, so prediction \cite{farazmand2019extreme} of the extreme events becomes a significant aspect for reducing the harmful loss \cite{herrera2014statistics}, or apply a suitable control scheme \cite{farazmand2019closed,ray2019intermittent,suresh2018influence} in advance. Hence the study on prediction of extreme events in the dynamical systems \cite{cavalcante2013predictability,kumarasamy2018extreme} or real-world scenarios deserve special attention. 
 
 \par For the prediction of extreme events in a dynamical system, it is essential to know the dynamics of the system in most cases. An analytical approach is available to detect an indicator for the prediction of the extreme events \cite{farazmand2016dynamical}, but it is a challenging task to apply this scheme irrespective of all systems. Bialonski et al.\ \cite{bialonski2015data} have proposed a data-driven prediction scheme for the prediction of extreme events in a spatially extended excitable system based on the reconstruction of the local dynamics from data. In the same time, machine learning models uncover a new scope of data-driven prediction of extreme events where it is not necessary to understand the nature of the system dynamics \cite{guth2019machine,amil2019machine,qi2020using,lellep2020using,pyragas2020using,narhi2018machine}. In the last few years, machine learning techniques are used extensively to explore several emergent phenomena in nonlinear dynamical systems. In this context, reservoir computing (RC) \cite{jaeger2004harnessing, maslennikov2019collective, shirin2019stability}, a version of recurrent neural network model, is effective for inference of unmeasured variables in chaotic systems using values of a known variable \cite{lu2017reservoir}, forecasting dynamics of chaotic oscillators \cite{pathak2018hybrid,fan2020long}, predicting the evolution of the phase of chaotic dynamics \cite{zhang2020predicting}, and prediction of critical transition in dynamical systems \cite{kong2021machine}. Also, RC is used to detect synchronization \cite{ibanez2018detection, weng2019synchronization, lymburn2019reservoir}, spiking-bursting phenomena \cite{saha2020predicting}, inferring network links \cite{banerjee2019using} in coupled systems.
 Apart from the RC, researchers have also applied different architectures of artificial neural networks such as feed-forward neural network (FFNN) \cite{svozil1997introduction, fine2006feedforward}, long-short term memory (LSTM) \cite{vlachas2020backpropagation, qin2019comparison, sangiorgio2020robustness} for different purposes such as detecting phase transition in complex network \cite{ni2019machine}, and functional connectivity in coupled systems \cite{frolov2019feed}, forecasting of complex dynamics \cite{vlachas2018data}. 
 

\par Currently, deep learning has furnished natural ways for humans to communicate with digital devices and is foundational for constructing artificial general intelligence \cite{sejnowski2020unreasonable}. Deep learning was persuaded by the architecture of the cerebral cortex and insights into autonomy and general intelligence may be found in other brain regions that are essential for planning and survival \cite{goodfellow2016deep}. Although applications of deep learning frameworks to real-world problems have become ubiquitous, but they are not without shortcomings: deep learners often exhibit high variance and may fall into local loss minima during training. At the forefront of machine learning and artificial intelligence, ensemble learning and deep learning have independently made a substantial impact on the field of time series forecasting through their widespread applications \cite{cao2020ensemble}. Ensemble learning combines several individual models to obtain better generalization of performance \cite{kuncheva2014combining}. Since the seminal paper by Bates and Granger \cite{bates1969combination}, various ensemble methods that combine different base forecasting methods have been introduced in the literature \cite{shaub2020fast, chakraborty2019forecasting, chakraborty2020real}. The so-obtained ensemble forecasting method is expected to have better accuracy than its components, but at the same time it should not overfit the data and should not be too complex to understand and explain. Thus, by creating an interpretable optimized ensemble deep learning model, one can utilize the relative advantages of both the deep learning models as well as the ensemble learning such that the final model has better generalization performance. Motivated by these, this paper proposes and studies a novel mathematical optimization-based ensemble deep learning model, namely optimized ensemble deep learning (OEDL) framework, to build an optimized ensemble which trades off the accuracy of the ensemble and utilizes the power of deep neural network models to be used. Our approach is flexible to incorporate desirable properties one may have on the ensemble, such as obtaining better predictive performance of the ensemble over individual forecasters. We illustrate our approach with real data sets arising in various contexts.

In our work, we endeavor to forecast the dynamics consisting of extreme events from the time-series using an optimized ensemble of three deep learning frameworks. 
Our main focus is to improve the forecasts given by three experts, namely (a) FFNN \cite{bebis1994feed}, (b) RC \cite{lukovsevivcius2012reservoir}, and (c) LSTM \cite{hochreiter1997long}. For this, we use the ensemble technique \cite{devaine2013forecasting} where three forecasts are combined using a best convex combination approach, and it gives a better result on prediction of regular events or extreme events than an individual one. The proposal should achieve synergistic improvements in model accuracy, stability, scalability, and reproducibility prompting a new wave of applications in the forecasting of dynamics. To validate our hypothesis, we experimentally show that it is efficient to use our proposed framework for getting better predictions than the individual experts (FFNN, RC, and LSTM). Besides applying the aggregation of forecasters on numerically simulated data sets, we also implement it for forecasting purposes of three real-world scenarios, like pandemic, epidemic, and weather event. The excellent performance of the OEDL framework in forecasting extreme events in Liénard-type system using synthetic data and also prediction of COVID-19 in Brazil, dengue cases in San Juan, and Ni{\~ n}o 3.4 ENSO prediction. In each case, we also compare the results obtained from the OEDL model with individual deep learners (FFNN, RC, and LSTM). The robustness and scalability of the proposed OEDL framework lie in its wide range of applications in various data sizes ranging from less than $500$ to $50000$ which makes it a potential forecaster for other applied forecasting problems. 

The main contributions of the paper can be summarized in the following manner:
\begin{enumerate}
    \item We present a novel formulation of the ensemble deep learning model consisting of FFNN, RC, and LSTM, based on a convex optimization problem that motivates our new forecasting method, optimized ensemble deep learning (OEDL). We also conclude that the theoretical formulation of the framework ensures worst-case guarantees and always generates the best set of weights (to be used in the ensemble) using an online solver.
    
    \item Using a range of tests with synthetic data collected from Liénard-type system consisting of extreme events comparing proposed OEDL against individual experts (FFNN, RC, LSTM) and simple ensemble method, we demonstrate that solving the predictability problem of extreme events to optimality yields an ensemble deep learner that better reflect the ground truth in the data.
    
    \item We extend the experimentation of the proposed OEDL framework to real-world forecasting challenges, for example, COVID-19 prediction in Brazil, dengue prediction in San Juan, and ENSO prediction in Ni{\~n}o 3.4 region, to show the wide range of applicability of the proposal.
    
    \item To practitioner's point of view, we present applications of the OEDL method on two synthetic data sets and three real-world data sets that predict with high accuracy when the optimal ensemble method will deliver consistent and significant accuracy improvements over individual experts. All the data sets and codes used in this study are made publicly available at \url{https://github.com/arnob-r/OEDL}.
\end{enumerate}

This article is arranged as follows. We introduce our optimized ensemble deep learning framework in Sec.\ \ref{ensemble}. In Sec.\ \ref{sec:result}, the comparison of forecasting accuracy among deep-learning forecasts and the ensemble of forecasts is the focused issue on both the numerically simulated and real-world data sets. We divide this section into two subsections. We describe the dynamics of the Li{\'e}nard-type system and discuss our result for numerically simulated data sets in Subsec.\ \ref{model}. Similar exploration has been done on real-world data sets in Subsec.\ \ref{real}, respectively. The conclusion is drawn in Sec.\ \ref{sec:Discussion} with some discussions and future aspects. In Appendix \ref{sec:forecasting}, we briefly explain three neural network-based forecasting methods to be used in the proposed ensemble framework.

\section{Optimized Ensemble Deep Learning Framework}\label{ensemble}
Deep learning is well known for its power to approximate almost
any function and increasingly demonstrates predictive accuracy
that surpasses human experts. However, deep learning models are
not without shortcomings; they often exhibit high variance and may
fall into local loss minima during training. Ensemble learning, as its name implies, combines multiple individual learners to complete the learning task together \cite{bates1969combination, wallis2011combining, shaub2020fast}. Indeed, empirical results of ensemble methods that combine the output of multiple deep learning models have been shown to achieve better generalizability than a single model \cite{ju2018relative}. In addition to simple ensemble approaches such as averaging output from individual models, combining heterogeneous models enables multifaceted abstraction of data, and may lead to better learning outcomes \cite{lee2015m}. Researchers' quest for an optimal solution to forecast combination continues in many applied fields ranging from dynamical systems to epidemiology. In this research, based on the constructed deep neural networks, an optimization algorithm is used to integrate the component learners for ensemble learning.

Section \ref{formulation} describes the formulation of the proposed OEDL framework in terms of an optimization problem with linear constraints. Section \ref{theoretical} establishes the connection of the approach with the constrained Lasso and some theoretical results of the solution are derived. Section \ref{algorithm} considers the algorithmic structure of the proposed OEDL framework. Finally, we discuss the implementation of the proposed OEDL model in Section \ref{implementation}.

\subsection{\label{formulation}Proposed OEDL: Model formulation}
This section presents the new ensemble approach. We describe the formulation of the model in terms of an optimization problem with linear constraints. Let $\mathcal{F}$ be a finite set of base forecasting models for the time series data $y_i$. No restriction is imposed on the collection of base forecasters. In this work, we introduce an OEDL framework which is an ensemble of three deep learning experts, namely FFNN, RC, and LSTM models.\\
By taking convex combinations of the base forecasters in $\mathcal{F}$, we obtain a broader class of forecasters, namely, 
\begin{equation}
\operatorname{op}(\mathcal{F})=\left\{F=\sum_{f \in \mathcal{F}} \alpha_{f} f: \sum_{f \in \mathcal{F}} \alpha_{f}=1, \alpha_{f} \geq 0, \; \forall \; f \in \mathcal{F}\right\}.
\label{opeqn}
\end{equation}
We denote $\left(\alpha_{f}\right)_{f \in \mathcal{F}}$ by ${\underline \alpha}$. The selection of an combined forecaster from $\operatorname{op}(\mathcal{F})$ can be done by optimizing a function which takes into account the following criteria. The fundamental criterion is the overall accuracy of the combined framework, measured through a loss function $\mathcal{L}$, defined on $\operatorname{op}(\mathcal{F})$,\\
$\begin{aligned} \mathcal{L}: \operatorname{op}(\mathcal{F}) & \longmapsto \mathbb{R}, \\ F & \longmapsto \mathcal{L}(F). \end{aligned}$ \\
Since in the proposed OEDL framework, we choose base forecasters $f \in \mathcal{F}$ with higher reliability, i.e., with lower individual loss, thus higher importance given to the loss of the ensemble forecaster. Therefore, an optimized ensemble learner is obtained by solving the following mathematical optimization problem with linear constraints:
\begin{equation}
\min_{{\underline \alpha} \in \mathrm{S}}\left\{\mathcal{L}\left(\sum_{f \in \mathcal{F}} \alpha_{f} f\right)\right\}
\label{eqn1}
\end{equation}
where $\mathrm{S}$ is the unit simplex in $\mathbb{R}^{|\mathcal{F}|}$,
\begin{equation*}
\mathrm{S}=\left\{{\underline \alpha} \in \mathbb{R}^{|\mathcal{F}|}: \sum_{f \in \mathcal{F}} \alpha_{f}=1, \alpha_{f} \geq 0, \; \forall \; f \in \mathcal{F}\right\}.
\end{equation*}

\subsection{\label{theoretical}Proposed OEDL: Theoretical results}
In general, Problem (\ref{eqn1}) has linear constraints. Based on the choice of loss functions, we can rewrite the objective function as a linear or a convex quadratic function while the constraints remain linear. Therefore, for the commonly used loss functions, Problem (\ref{eqn1}) is easily tractable with commercial solvers. In addition, under some mild assumptions, we can characterize the behavior of the optimal solution.
\begin{remark}
Let $T$ be a training sample in which each individual time sequences $y_i \in T$. Suppose $\mathcal{L}$ be the empirical loss of quantile regression for $T$, i.e.,
$$\mathcal{L}\left(\sum_{f \in \mathcal{F}} \alpha_{f} f\right)=\sum_{i \in T} \rho_{\tau}\left(y_{i}-\sum_{f \in \mathcal{F}} \alpha_{f} f\right),$$
where
$$\rho_{\tau}(k)=\left\{\begin{array}{ll}\tau k, & \text { if } \; k \geq 0 \\ -(1-\tau) k, & \text { if } \; k<0\end{array}\right.$$
for some $\tau \in(0,1)$. Then, Problem (\ref{eqn1}) can be expressed as a linear programming problem and thus efficiently solved with linear programming solvers \cite{koenker2001quantile, koenker2005inequality}.
\end{remark}
\begin{remark}\label{rem2}
Let $T$ be a training sample in which each individual time sequences $y_i \in T$. Suppose $\mathcal{L}$ be the empirical loss of ordinary least squares  regression for $T$, i.e.,
$$\mathcal{L}\left(\sum_{f \in \mathcal{F}} \alpha_{f} f\right)=\sum_{i \in T}\left(y_{i}-\sum_{f \in \mathcal{F}} \alpha_{f} f\right)^{2}.$$
Hence, Problem (\ref{eqn1}) is a convex quadratic optimization problem with linear constraints, can be seen as a constrained Lasso problem \cite{gaines2018algorithms} without standard regularization parameter ($\lambda = 0$). In particular, we can assert that Problem (\ref{eqn1}) has unique optimal solution ${\underline \alpha}$.
\end{remark}
Under mild conditions on $\mathcal{L}$, applicable in particular for the quantile and ordinary least squares empirical loss functions, we can find the optimal solution of the Problem (\ref{eqn1}). For the quantile and ordinary least squares empirical loss functions, these constraints are either linear or convex quadratic, and thus the optimization problems can be addressed with the available convex solvers \cite{cesa2006prediction}. However, we are interested in online aggregation rules that perform almost as well as, for instance, the best constant convex combination of the experts to solve Problem (\ref{eqn1}). In our proposed OEDL framework, we use \textit{`oracle'} function \cite{cesa2006prediction, devaine2013forecasting, gaillard2015forecasting} which is the best constant convex combination of the experts; in fact, they hold for all sequences of time series and come with finite time worst-case guarantees.

\subsection{\label{algorithm}Proposed OEDL: Algorithm and flow diagram}
\begin{figure}
	\includegraphics[scale=0.50]{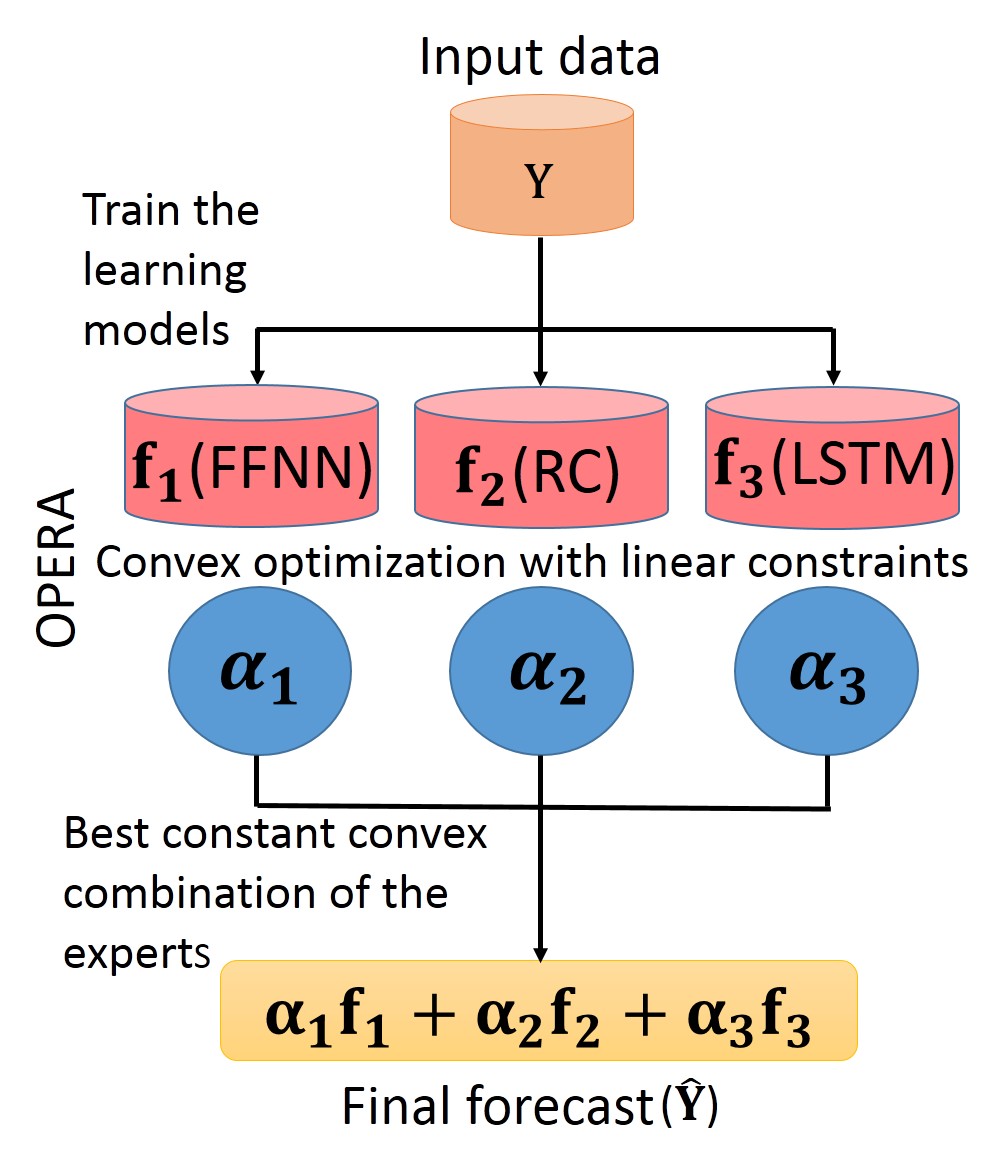}%
	\caption{The schematic diagram depicts the aggregation of expert forecasters. For a given input vector, three deep learning frameworks FFNN, RC, and LSTM generate three sets of forecasts. We combine the three results and get a new forecast with better accuracy with respect to RMSE performance. The aggregation rule is defined as a best constant convex combination of forecasts. For $i$-th time if FFNN, RC, and LSTM produce the forecast values $f_1$, $f_2$, and $f_3$, then the final forecast is $\alpha_1 f_1+\alpha_2 f_2+\alpha_3 f_3$ subject to $\alpha_1+\alpha_2+\alpha_3=1 \; (\alpha_i \geq 0)$.}
	\label{figure_5} 
\end{figure}

In the proposed OEDL framework, we create combination (ensemble) of forecasts based on optimized online expert aggregation method from a pool of forecasting methods, namely 
\begin{enumerate}
    \item Feed-forward neural network (Appendix \ref{sec:forecasting}A);
    \item Reservoir computing (Appendix \ref{sec:forecasting}B);
    \item Long short-term memory (Appendix \ref{sec:forecasting}C).
\end{enumerate}
More formally, we consider a sequence of observations $y_1,y_2,\ldots,y_n$ (any real bounded time series) to be predicted step by step. Suppose that a finite set of deep learning-based forecasters $k= 1,2,3$ (FFNN, RC, and LSTM) provide us before each time step $t=1,2, \ldots, n$ predictions $f_{k, t}$ of the next observation $y_{t}$. We obtain the final prediction $\hat{y}_{t}$ by using only the knowledge of the past observations $y_1,y_2,\ldots,y_{t-1}$ and expert forecasts $f_{1, t},f_{2, t}, f_{3,t}$, i.e.,
\begin{equation}\label{result_final}
\hat{y}_{t}=\sum_{k=1}^{3} \alpha_{k, t} f_{k, t},
\end{equation}
where $\alpha_{1,t}$, $\alpha_{2,t}$ and $\alpha_{3,t}$ are three non-negative weights subject to $\alpha_{1,t}+\alpha_{2,t}+\alpha_{3,t}=1$ as in Eqn. (\ref{opeqn}).
Interestingly, we can choose these weights equal and uniformly \cite{shaub2020fast} in such a way that $\alpha_{1,t}=\alpha_{2,t}=\alpha_{3,t}=\frac{1}{3}$ (we refer it as `ensemble' model) but this formulation does not come with finite time worst-case guarantees. We formulate this mathematical optimization problem in terms of an empirical squared loss minimization problem as mentioned in Remark \ref{rem2} with the following loss function,
\begin{equation}\label{final_eqn}
\mathcal{L}\left(\sum_{k=1}^{3} \alpha_{k,t} f_{k,t}\right)=\sum_{t=1}^{n}\left(y_{t}-\sum_{k=1}^{3} \alpha_{k,t} f_{k,t}\right)^{2}.
\end{equation}
We obtain the best possible weights in the proposed OEDL framework by minimizing the above loss function. This choice of loss function ensures that the best convex oracle strategy by an efficient aggregation performance of forecasters can be obtained in comparison to individual experts or the uniform average of the expert forecasts with respect to mean squared error. The function oracle performs a strategy that cannot be defined online and requires in advance the knowledge of the whole data set $y_t$ and the expert advice to be well defined. {\it `Online Prediction by ExpeRts Aggregation' (opera)} is a robust online solver to estimate these weights $(\alpha_{k,t})$ based on expert forecasts, developed by Gaillard et al.\ \cite{devaine2013forecasting, gaillard2015forecasting}. The online solver \textit{opera} performs, for regression-oriented time-series, predictions by combining a finite set of forecasts provided by the user (FFNN, RC, and LSTM in our case). To check the accuracy of prediction, we select the following statistical metric, namely root mean square error (RMSE) \cite{klos2020dynamical} as defined below,
\begin{equation}\label{eq.3}	
RMSE=\sqrt{\dfrac{1}{n} \displaystyle \sum_{t=1}^{n} (y_t-\hat y_t)^2}.
\end{equation}
The complete procedure for making one-step-ahead predictions with our optimized ensemble methodology is summarized in Algorithm \ref{algo1} and can be visualized in Fig. \ref{figure_5}.
\begin{algorithm}\label{algo1}
Offline phase: Train the learning models (FFNN, RC and LSTM).\\
\KwIn{$\{y_1, y_2,\ldots,y_n\}$: $n$ observed time series that form the reference set.\\
$\mathcal F$: a set of expert forecasts to be used in the ensemble.}
\KwOut{Record $\{\hat y_t: t=1,2,\ldots,n\}$}
\textbf{Steps:}\\
\nl  . For each family of strategies (introduced in Eqn. \ref{opeqn}) compute the performance corresponding to the best constant choices of the parameters by minimizing Eqn. \ref{final_eqn}. \\
\nl . Obtain the best set of weights ($\alpha_f$) using the \textit{`oracle'} function (set of convex weights) available in the online linear time solver, namely {\it `Online Prediction by ExpeRts Aggregation' (opera)}.\\
\nl . Obtain the final set of predictions from the OEDL model using Eqn. \ref{result_final}.\\
\nl . Assess the quality of the operational performance, i.e., the performance obtained after some automatic and sequential tuning.\\
\nl . Compare the performance of the experts as well as uniform combination of experts and best convex combination of experts as in OEDL using RMSE as in Eqn. \ref{eq.3} (on new data).
\caption{{\bf Optimized ensemble deep learning}
}
\end{algorithm}

\begin{figure}[t]
	\includegraphics[scale=0.25]{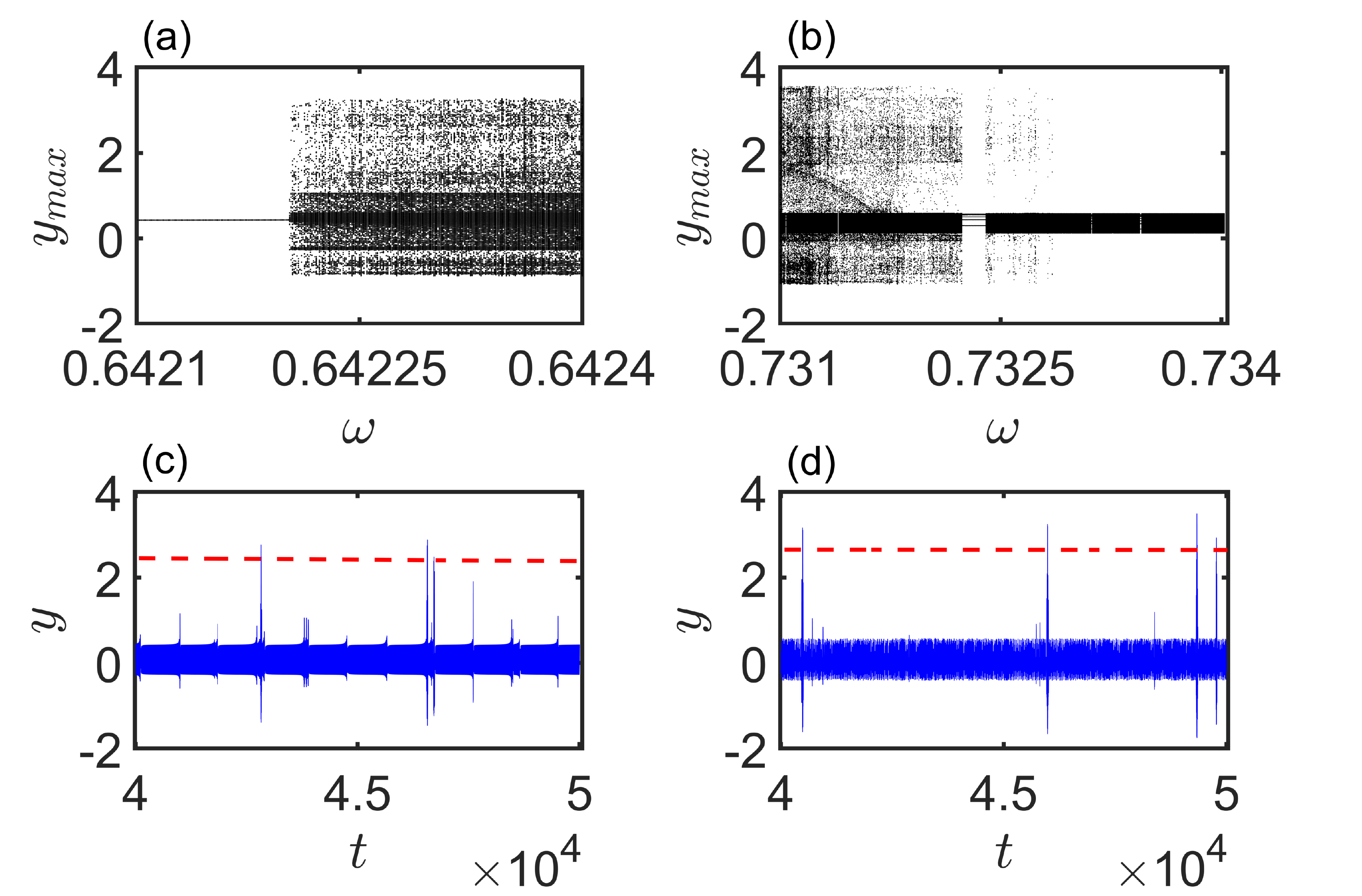}%
	\caption{Bifurcation diagram in the upper panel. Variation of local maxima of $y$ ($y_{max}$) by varying the parameter $\omega$ is depicted for (a) $\omega\in[0.6421,0.6424]$ and (b) $\omega\in[0.731,0.734]$. A sudden transition take places for both the cases. For sub-figure (a), periodic attractor transits to chaotic attractor at $\omega\approx0.6422$ and chaotic attractor appears via intermittency. In (b), if we decrease the $\omega$, then size of chaotic attractor suddenly expands at $\omega\approx0.7328$ due to interior crisis. After crossing the critical value of $\omega$ in the suitable direction, a range of $\omega$ can be found, where system exhibits chaos with intermittent high amplitude oscillation in the temporal dynamics. Time series of $y$ in the lower panel. Temporal evolution of $y$ at (c) $\omega=0.6423$, and sub-figure (d) $\omega=0.7315$. Both figures display extreme events as few values of $y_{max}$ exceed the extreme event qualifying threshold, $H_S$ (dashed horizontal line). Other parameters: $\alpha'=0.45, \beta'=0.5, \gamma'=-0.5,$ and $F'=0.2$.}
	\label{figure_1} 
\end{figure}

\begin{figure}
	\centerline{
		\includegraphics[scale=0.65]{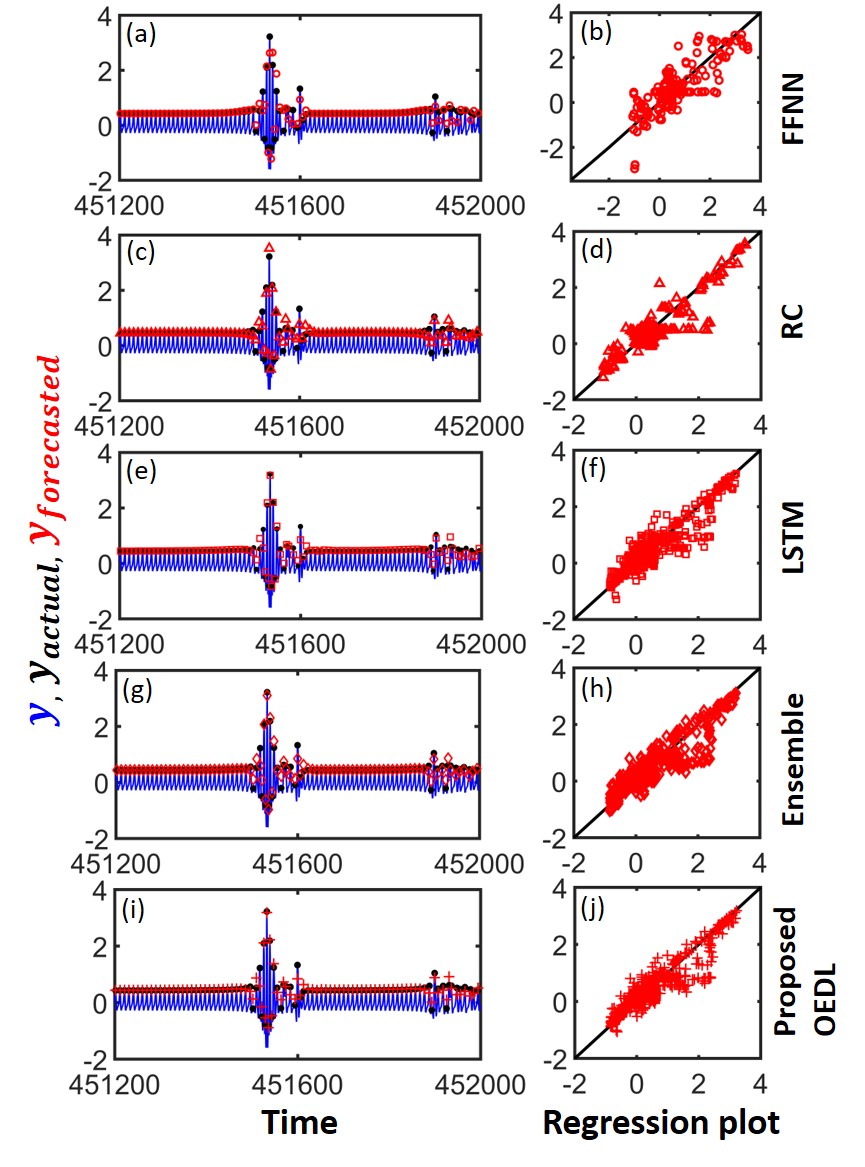}}
	\caption{Forecasting of events in the left panel. The short segment of the time evolution of $y$ (blue line) is drawn for $\omega=0.6423$ in each sub-figure. FFNN, RC, and LSTM are trained for $45000$ time points of the data set, and the rest of the data are used for forecasting. The events (local maxima of those time series and denoted by black dots), lying on the test data sets, are forecasted by the five forecasters. Forecasting values are denoted by the circle (for FFNN), triangle (for RC), square (for LSTM), diamond (ensemble of experts), and plus (proposed OEDL). Regression plot in the right panel. The scatter plots are illustrated in the right panel using test events along $x$-axis and their predictions along $y$-axis for each forecaster.}
	\label{figure_6a} 
\end{figure}

\subsection{Proposed OEDL: Model implementation \label{implementation}}
Proposed OEDL refers to an optimized ensemble model where instead of
building a single model, multiple `base' models are combined to perform
time series forecasting tasks. The implementation of the proposed OEDL framework consists of four major steps:\\
(a) Time-series data is divided into training and test data sets. Neural network models (FFNN, RC, and LSTM) are implemented on the training data sets. \\
(b) Obtain the forecasts based on these experts (on the test data) and use them in the next step.\\
(c) Generating the final ensemble forecasts using the best constant combination of the experts using \textit{Online Prediction by ExpeRts Aggregation}. \\
(d) Compare the predicted test results of proposed OEDL with individual expert forecasts and ensemble (with uniform weights) based on a statistical measure (RMSE). 

Now, we discuss the practical implementation details of individual experts and ensemble methods to be used in this paper. FFNN model is built using publicly available \href{http://matlab.izmiran.ru/help/toolbox/nnet/newff.html}{\color{blue}Neural Network Toolbox} in MATLAB while for LSTM model we use MATLAB library \href{https://in.mathworks.com/help/deeplearning/ug/time-series-forecasting-using-deep-learning.html}{\color{blue}Time Series Forecasting Using Deep Learning}. For RC model implementation, we have used the MATLAB implementation of the \href{https://github.com/stefanonardo/echo-state-network}{\color{blue}ESN} function. To build the ensemble model with equal (uniform) weights we take a simple arithmetic average of the expert forecasts and generate the results for the basic `ensemble' model. An optimized ensemble deep learning model is built using the available forecasts based on the three deep neural network-based experts. To obtain the best convex combination of the experts, we have used an online mathematical optimization solver, namely \href{https://cran.r-project.org/web/packages/opera/vignettes/opera-vignette.html}{\color{blue}{OPERA}}. Using the online solver, we obtain the best combination of weights for the proposed OEDL framework. For the sake of reproducibility, the data sets and implementation code of the proposed OEDL framework are made publicly available in our \href{https://github.com/arnob-r/OEDL}{\color{blue}GitHub} repository. 

\section{\label{sec:result}Applications}
In this section, we implement the proposed OEDL framework on both numerically simulated and real-world data sets. We divide this section into two subsections. Firstly, we describe the paradigmatic model and then discuss our result on Li{\'e}nard-type system. Secondly, we experiment with the proposed methodology on three real-world data sets. Each data set (numerical as well as real-world) consists of extreme events from the perspective of each situation.  

\subsection{\label{model}The Li{\'e}nard-type system}
\par We consider a paradigmatic Li{\'e}nard-type system \cite{chandrasekar2005unusual} with an external sinusoidal forcing. Leo Kingston et al.\ \cite{kingston2017extreme} have reported that this system exhibits extreme events for an admissible set of parameter values. The mathematical expression of the model is presented by 
\begin{equation}
\begin{array}{l}\label{eq.1}	
\dot{x} =y,\\
\dot{y} =-\alpha' xy-\gamma' x-\beta' x^3+F' sin(\omega t),\\
\end{array}
\end{equation}
where $\alpha', \beta'$ and $\gamma'$ represent nonlinear damping,  strength of nonlinearity and intrinsic frequency, respectively. $F'$ and $\omega$ are the amplitude and frequency of the external forcing, respectively. The values of the parameters are kept fixed at $\alpha'=0.45, \; \beta'=0.5, \; \gamma'=-0.5,$ and $F'=0.2$. The parameter $\omega$ is treated as a bifurcation parameter, and it is varied to observe extreme events in this system. Leo Kingston et al.\ \cite{kingston2017extreme} 
have showed that extreme events occurs in $y$ variable of this system via Pomeau-Manville intermittency \cite{pomeau1980intermittent} and interior-crisis induced intermittency \cite{grebogi1987critical} route to chaos. So, the state variable $y$ is observable in this system and the local maxima are considered as events. Two bifurcation diagrams are plotted to reveal two routes in the upper panel of Fig.\ \ref{figure_1}. The local maxima of $y$, $y_{max}$ are portrayed by varying the forcing frequency parameter $\omega \in [0.6421,0.6424]$ in Fig.\ \ref{figure_1}(a), and $\omega \in [0.731,0.0.734]$ in Fig.\ \ref{figure_1}(b). In Fig.\ \ref{figure_1}(a), when the value of $\omega$ is increased, we observe that period-1 orbit suddenly transits to chaotic attractor at a critical value of $\omega$ ($\omega\approx0.6422$). This scenario is called Pomeau-Manville intermittency \cite{pomeau1980intermittent}. On the other hand, the size of the chaotic attractor suddenly decreases when the value of $\omega$ crosses the critical value of $\omega$ ($\omega\approx0.7328$) in Fig.\ \ref{figure_1}(b). Here, high amplitude chaotic attractor transfers to the chaotic attractor with different low amplitude via interior crisis. Due to the interior crisis, intermittent chaotic bursts are observed in the time series of $y$ at the vicinity of the critical value of $\omega$. An event ($y_{max}$) is called extreme event \cite{kingston2017extreme} if it occasionally crosses a significant height, $H_{S}=\mu+8\sigma$ \cite{zanna2008oceanic,kharif2008rogue,chowdhury2020distance}, a pre-defined threshold. Here $\mu$ and $\sigma$ are mean and standard deviation of a sufficiently large data set of events, respectively. 
We select two values of $\omega$ from the bifurcation diagram Figs.\ \ref{figure_1}(a)-(b) so that the chaotic attractors corresponding to the values of $\omega$, exhibiting extreme events, are manifested through the different routes, as mentioned above. We plot the time evolution of $y$ in  Fig.\ \ref{figure_1}(c) at $\omega=0.6423$, and in Fig.\ \ref{figure_1}(d) at $\omega=0.7315$, respectively. Extreme events are noticed from both figures as few high amplitude values of $y_{max}$ exceed $H_S$ (red dashed line in Figs.\ \ref{figure_1}(c)-(d)). For prediction purpose, we construct two data sets of $y_{max}$ for $\omega=0.6423$, and $\omega=0.7315$.

\subsubsection{Training and prediction in Li{\'e}nard-type system}
We consider two data sets of $y_{max}$ from Li{\'e}nard-type system via simulating Eqn.\ \eqref{eq.1}, constructed for $\omega=0.6423$, and $\omega=0.7315$. Two mentioned values of $\omega$ are chosen in such a way that the chaotic attractor is generated due to Pomeau-Manville intermittency, and interior crisis route to chaos, respectively. Each data set consists of $50000$ events. Out of $50000$ data points, $90\%$ of events are trained to the FFNN, RC, and LSTM. Rest $10\%$ cases are used to test the forecasting results. For the experimentation on numerically simulated data sets, the hidden layer in the FFNN model consists of $100$ neurons and the learning rate for training data is $0.05$. For RC, the size of the reservoir is $400$, and the average degree of $W_{res}$ is $40$. The number of hidden cells is chosen as $200$ in the LSTM layer. More precisely, the FFNN model is trained for $2500$ epochs. while for training the LSTM model, the data points are trained for $250$ epochs. The gradient threshold is $1$ to prevent the gradients from exploding. The initial learning rate $0.005$ and drop the learning rate after $125$ epochs by multiplying by a factor of $0.2$. 

\begin{figure}
	\centerline{
		\includegraphics[scale=0.65]{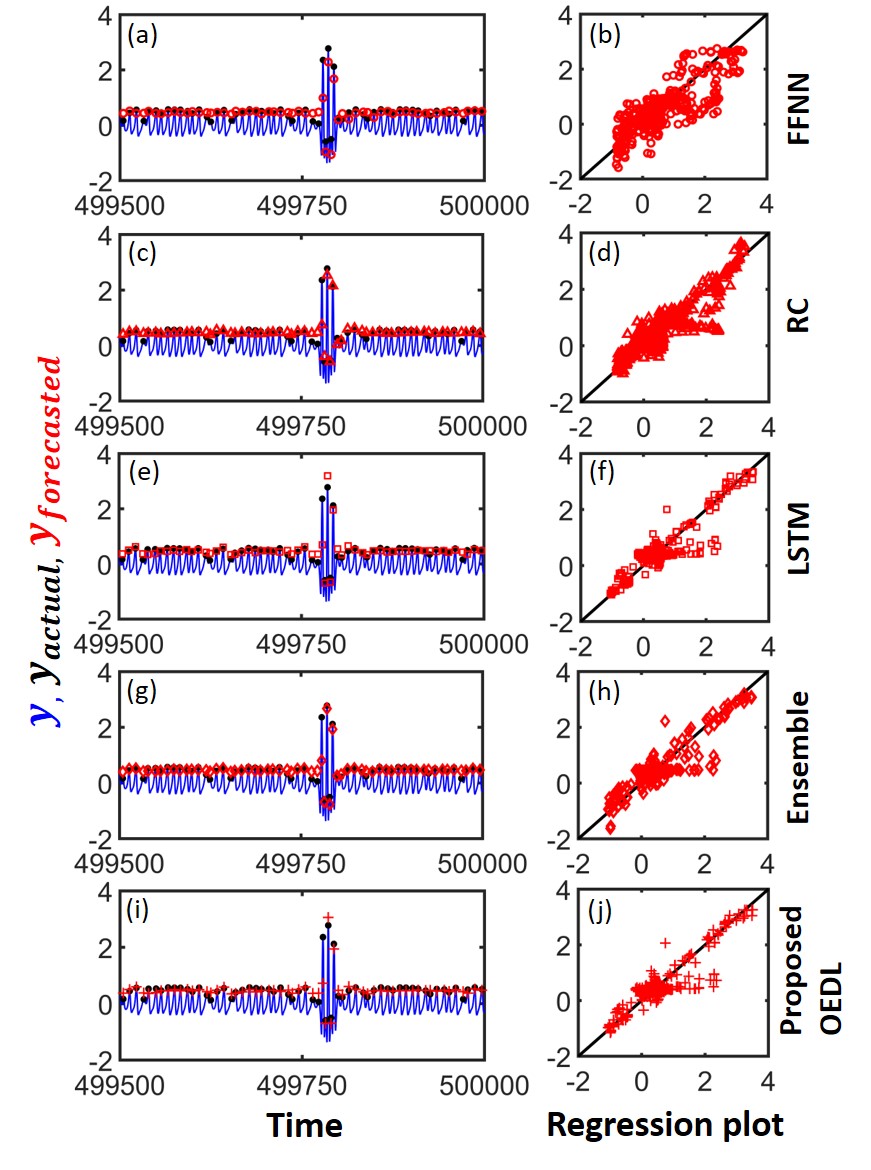}}
	\caption{Forecasting of events in the left panel. The short segment of the time evolution of $y$ (blue line), events (black dots), and corresponding forecasted values are drawn for $\omega=0.7315$ for each forecaster in each sub-figure. Forecasting values are denoted by the circle (for FFNN), triangle (for RC), square (for LSTM), diamond (ensemble of experts), and plus (proposed OEDL).  Regression plot in the right panel. The scatter plots are plotted using test events along the $x$-axis and their predictions along the $y$-axis for each forecaster. All the models are trained for $45000$ time points and the rest event points are used for testing.}
	\label{figure_6b} 
\end{figure}

\subsubsection{Results}
We aggregate three forecast outputs of individual FFNN, RC, and LSTM models using two aggregation rules. Firstly, we implement the usual ensemble technique which is an average (arithmetic mean) of three expert forecasts. For standard ensemble deep learning model \cite{bates1969combination, shaub2020fast}, $\alpha_1=\alpha_2=\alpha_3=\frac{1}{3}$. The results on test data sets are computed for ensemble framework and average RMSE and its standard deviations are shown in Table \ref{table1}. 

\par Now, we apply our proposed OEDL model to these two simulated data sets by following the algorithm \ref{algo1} as discussed in Subsec.\ \ref{algorithm}. In the left panel of Fig.\ \ref{figure_6a}, we plot a small segment of the time evolution of $y$-variable (blue) consisting of extreme events, and the events ($y_{actual}$) are marked by black dots for the data set of $\omega=0.6423$.  For this data set corresponding to $\omega=0.6423$ from Li{\'e}nard-type system, we calculate the weights $\alpha_1= 0.144, \alpha_2=0.133, \alpha_3=0.723$ for convex aggregation rule. In Figs.\ \ref{figure_6a} (b, d, f, h, j), we draw the regression plots corresponding to Figs.\ \ref{figure_6a} (a, c, e, g, i), respectively. Similarly, time evolution of $y$ along with the events ($y_{actual}$), and forecasting values ($y_{forecasted}$) are depicted for $\omega=0.7315$ in the left panel of Fig.\ \ref{figure_6b}. Also, corresponding respective regression plots are exhibited in the right panel of Fig.\ \ref{figure_6b}. We determine $\alpha_1=0.101, \alpha_2=0.066, \alpha_3=0.833$ for the second data set corresponding to $\omega=0.7315$. We show regression pots using test data points of $y_{max}$ and data sets of forecasts using different frameworks and aggregation rules in the left panel of Figs.\ \ref{figure_6a} (for $\omega=0.6423$) and \ref{figure_6b} (for $\omega=0.7315$), respectively. Figures\ \ref{figure_6a} and \ \ref{figure_6b} give visual comparisons of the performances by different forecasters including proposed OEDL model. 
\begin{table}
\centering
\caption{Experimental results (values are calculated after 30 runs) for individual forecasting models (FFNN, RC, LSTM), Ensemble model and proposed OEDL model for two data sets from Li{\'e}nard-type system. $\langle \mbox{RMSE} \rangle$ indicates the mean RMSE values and $\sigma_{\mbox{RMSE}}$ denotes the standard deviations. The best result is obtained using the proposed OEDL model and shown in the last row (bold).}
\begin{tabular}{|l|c|c|c|c|}
\hline  & \multicolumn{2}{|c|} {$\omega=0.6423$} & \multicolumn{2}{c|} {$\omega=0.7315$} \\
\cline { 2 - 5 } & $\langle \mbox{RMSE} \rangle$ & $\sigma_{\mbox{RMSE}}$ & $\langle \mbox{RMSE} \rangle$ & $\sigma_{\mbox{RMSE}}$ \\
 & & & & \\ \hline
FFNN & 0.1914 & 0.0055 & 0.1724 & 0.0043 \\
RC & 0.1595 & 0.0007 & 0.1488 & 0.0003 \\
LSTM & 0.1533 & 0.0050 & 0.1273 & 0.0065 \\
Ensemble & 0.1507 & 0.0024 & 0.1369 & 0.0028 \\
{\color{blue}Proposed OEDL} & \textbf{0.1483} & 0.0029 & \textbf{0.1266} & 0.0062 \\ \hline
\end{tabular}\label{table1}
\end{table}

Once we get the weights for each of the expert models, these are put in Eqn.\ \eqref{result_final} to obtain the final forecasts for the proposed OEDL model. Finally, using Eqn.\ \eqref{eq.3}, we calculate RMSE values for our proposal and get a clear picture regarding the significant improvement in the predictive performance of the OEDL model over FFNN, RC, LSTM, and simple ensemble models. We repeat the processes for $30$ times to check the robustness of our result and report the average and standard deviations of RMSE values in Table \ref{table1} for all the five models considered in this study.
For both cases, we observe that the proposed OEDL model gives the best result according to the RMSE (minimum RMSE values). This is because we minimized the loss function corresponding to convex aggregation as discussed in Sec.\ \ref{ensemble} and the success of the proposal is also evident from both the Figs.\ \ref{figure_6a} and \ref{figure_6b}. 


\begin{figure*}[ht]
	\includegraphics[scale=0.78]{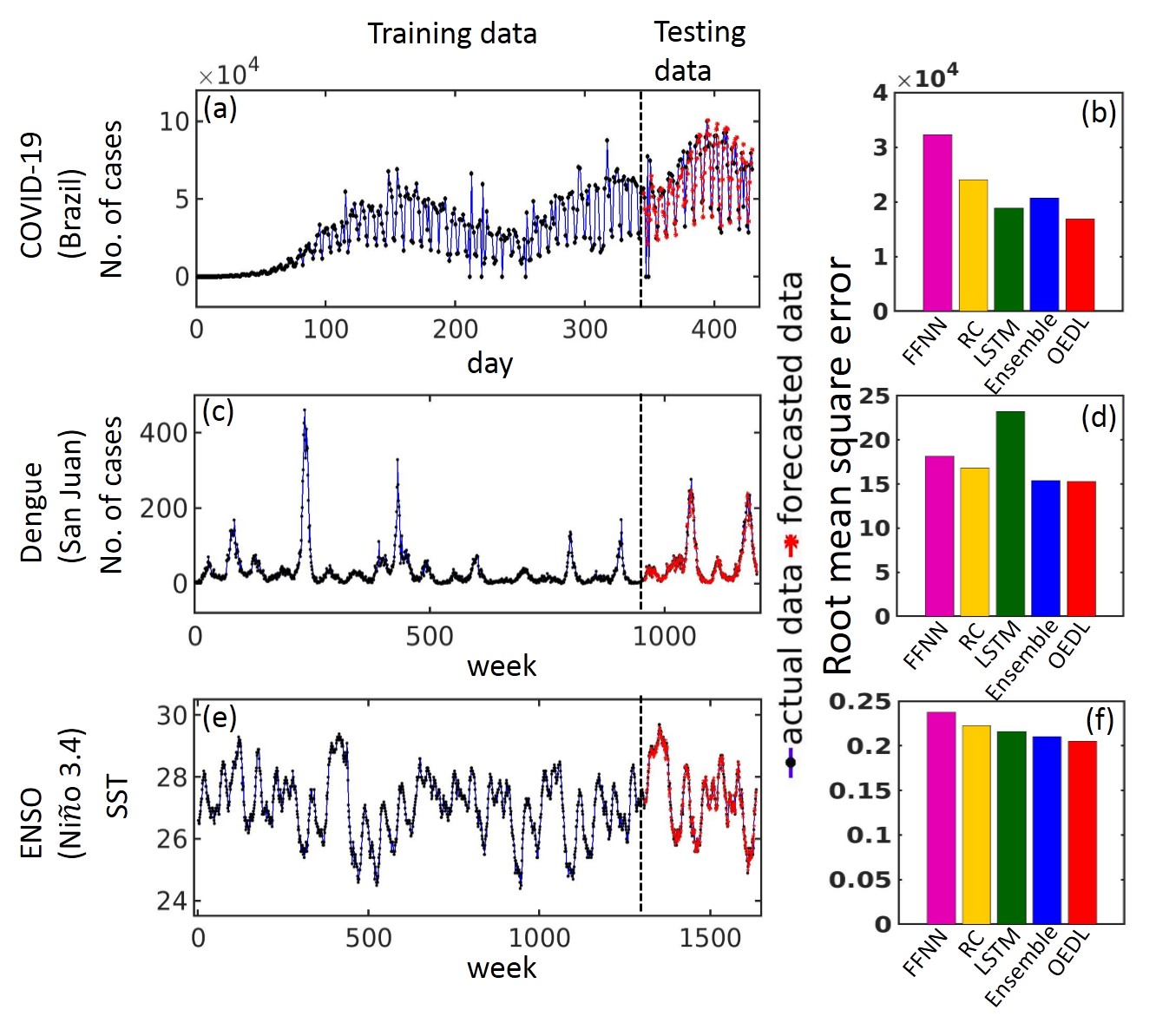}%
	\caption{(a, c, e) Actual vs predicted forecasts: A comparison between actual data points (black dots) and forecasted data points (red star) is shown in the left panel for three real-world cases, (a) COVID-19 forecasting, (c) dengue forecasting, and (e) prediction of sea surface temperature. For each case, the training and testing phase are separated by the black dashed vertical line. Only forecasting results by the proposed OEDL method is depicted in (a, c, e). (b, d, f) The values of root mean square function corresponding to five forecasters are presented by bar plots in the right panel for the mentioned real-world scenarios. The values of RMSE are minimum corresponding to the proposal across all the data sets. So, the proposal does considerably well for giving better accuracy than other forecasting methods.}
	\label{figure_9} 
\end{figure*}

\subsection{Real-world data sets\label{real}}
Now, we validate the proposed OEDL model for real-world data and check whether it gives a better prediction or not. For this, we experiment on various real-world data sets of different dynamics. Three time-series forecasting data sets from practical fields are considered to showcase the scalability of the proposal. These data sets are relatively small in size in comparison to numerically simulated data. First, we take data of daily cases of the recent COVID-19 pandemic in Brazil. Another data set consists of weekly dengue cases at San Juan, capital of Puerto Rico. Finally, one more data set of sea surface temperature (SST) values in the Ni{\~ n}o 3.4 regions is also considered for our study. We use $80\%$ of the data points for training the forecasting methods. Rest $20\%$ of the data sets is used for checking the performance of the forecasters. All these data sets contain extreme events from the perception of the research fields. Below we provide a brief description of these real-world data sets.

\subsubsection{Description of data sets}
(a)\textit{ COVID-19 pandemic data:} A daily confirmed cases of COVID-19 from $26$th February, $2020$ to $29$th April, $2021$ in Brazil are collected from \href{https://ourworldindata.org/coronavirus-source-data}{\color{blue}Our World in Data}. The data set consists of $429$ data points in total. Several works have been done regarding COVID-19 forecasting \cite{chakraborty2020real, petropoulos2020forecasting, perc2020forecasting}. From the Brazil pandemic data set, we take $80\%$ of the total data (i.e., $343$ daily cases) as training data and rest as test data set.\\

(b) \textit{Dengue epidemic data:} We gather $1196$ weekly cases of dengue from May, $1990$ to April, $2013$ of San Juan from the \href{https://dengueforecasting.noaa.gov/}{\color{blue}National Oceanic and Atmospheric Administration}. Dengue forecasting is one of the current research fields of epidemic forecasting \cite{baquero2018dengue, chakraborty2019forecasting, benedum2020weekly}. We build our models on $956$ weekly cases ($80\%$ of total data) for the training phase and the rest for testing purpose in our study.\\

(c) \textit{ENSO data:} El Ni{\~n}o-Southern Oscillation (ENSO) is a climate phenomenon that occurs due to an interplay between atmospheric and ocean circulations \cite{allan1996nino}. A fluctuation in sea surface temperature (SST) is observed across the equatorial Pacific Ocean. A warming phase of SST in the eastern is called El Ni{\~n}o, and La Ni{\~n}a is an occasional cooling of ocean surface waters in the eastern Pacific Ocean \cite{ray2020enso}. These two phases are captured through the data set consisting of sea surface temperatures of the Pacific Ocean. We collect weekly sea surface temperatures from $3$rd January, $1990$ to $21$st April, $2021$ in the Ni{\~ n}o 3.4 region ($5^{\circ}$ North-$5^{\circ}$ South and $170^{\circ}-120^{\circ}$ West) across the Pacific Ocean from the \href{https://www.cpc.ncep.noaa.gov/data/indices/}{\color{blue}Climate Prediction Center} to form a set of $1634$ data points. This data can be used for SST forecasting or to say El Ni{\~ n}o / La Ni{\~ n}a forecasting. Prediction of El Ni\~no is one of the trending topics in the field of climatology \cite{chen2004predictability, dijkstra2019application}. We also validate our proposed framework for forecasting SST over Ni{\~ n}o 3.4 regions, and it helps for Ni{\~ n}o 3.4 ENSO prediction. We train $1307$ data points of SST (weekly collected) and the rest data points are used for testing purposes in this case study.

\subsubsection{Results on real-world data sets}
The standard implementation of FFNN, RC, and LSTM methods is followed for all the real-world data sets as discussed in Section \ref{implementation}. The number of hidden neurons in FFNN, size of the reservoir in RC, number of hidden cells in LSTM are chosen using the cross-validation technique \cite{james2013introduction, goodfellow2016deep}.


Now, we explore the forecasting results of real-world data sets using five different forecasters to showcase the excellent performance of our proposed OEDL model (convex combination of experts) in terms of RMSE. In the left panel of Fig.\ \ref{figure_9}, the time series (blue) is displayed where Figs.\ \ref{figure_9}(a), \ref{figure_9}(c), and \ref{figure_9}(e) depict the time evolution of the daily cases of COVID-19, the weekly cases of dengue, and the weekly record of SST over Ni{\~n}o 3.4 regions. The right-hand side of a black dashed line of each mentioned sub-figures indicates the training data points (black dots). The left-hand side of that dashed line demonstrates the out-of-sample forecasting data points (red star) against actual data points. Here, we only show the forecasting results given by the proposed convex combination of three deep learning experts (FFNN, RC, LSTM). The comparative study between actual and forecasted data points is illustrated from those figures. Besides, the RMSE values for five competitive forecasting methods are given in the right panel of Fig.\ \ref{figure_9} for three different real-world data sets. The lesser the value of RMSE, the better the forecasting method is. Five bars are plotted at Figs.\ \ref{figure_9}(b, d, f) corresponding to five forecasters, namely FFNN (magenta), RC (yellow), LSTM (green), and basic ensemble method (blue) and proposed OEDL method (red), respectively. According to three Figs.\ \ref{figure_9}(b, d, f), we observed that the proposed OEDL attains minimum RMSE among other methods for all the three real-world data sets. Let forecasting results of FFNN, RC, and LSTM are aggregated with the following weights $\alpha_1$, $\alpha_2$, and $\alpha_3$, respectively in the proposed framework. So, for the chosen sets of weights corresponding to three real-world case studies are given here, and are used for depicting the Fig.\ \ref{figure_9}. We found $\{\alpha_1, \alpha_2, \alpha_3\}=\{0.0, 0.33, 0.67\}$ for the case of daily cases of COVID-19 forecasting, $\{\alpha_1, \alpha_2, \alpha_3\}=\{0.31, 0.425, 0.265\}$ for weekly dengue cases forecasting, and $\{\alpha_1,\alpha_2,\alpha_3\}=\{0.029, 0.41, 0.561\}$ for SST forecasting. Also, for this real-world data sets, we compute the mean ($\langle \mbox{RMSE} \rangle$) and fluctuation ($\sigma_{\mbox{RMSE}}$) of the RMSEs for each forecasting methods in the Table\ \ref{table2} to check the practical robustness of our proposal. Here, the values of RMSE are calculated after $50$ trials for all the three real-world data sets. From Table\ \ref{table2}, we conclude that the proposed OEDL always wins from the perspective of forecasting accuracy in comparison with the results of other experts or uniform aggregation of experts (ensemble) in a significant margin. 
\begin{table}[!ht]
\centering
\caption{Mean RMSE ($\langle \mbox{RMSE} \rangle$) and standard deviation ($\sigma_{\mbox{RMSE}}$) of RMSEs which are evaluated over $50$ realizations (on test data sets) for FFNN, RC, LSTM, ensemble and proposed OEDL are given here. The left panel, middle panel, and right panel delineate for the case of COVID-19 forecasting, dengue forecasting, and prediction of SST, respectively. The best results are made bold in the last row.}
\makebox[0.8\width]{
\begin{tabular}{|l|c|c|c|c|c|c|}
\hline  & \multicolumn{2}{|c|} { COVID-19 (Brazil) } & \multicolumn{2}{c|} { Dengue (San Juan) } & \multicolumn{2}{|c|} { ENSO (Nino 3.4) } \\
\cline { 2 - 7 } & $\langle \mbox{RMSE} \rangle$ & $\sigma_{\text {RMSE }}$ & $\langle \mbox{RMSE} \rangle$ & $\sigma_{\text {RMSE }}$ & $\langle \mbox{RMSE} \rangle$ & $\sigma_{\text {RMSE }}$ \\
& & & & & & \\ \hline
\hline 
FFNN & $31403.8848$ & $1781.1926$ & $18.0829$ & $0.2737$ & $0.2571$ & $0.0401$ \\
RC & $24075.1758$ & $145.8909$ & $16.8062$ & $0.2205$ & $0.2232$ & $0.0005$ \\
LSTM & $16387.1348$ & $1667.7996$ & $23.7285$ & $3.0416$ & $0.2093$ & $0.0057$ \\
Ensemble & $21928.0723$ & $869.6142$ & $16.6957$ & $0.6287$ & $0.2164$ & $0.0058$ \\
{\textbf{OEDL}} & $\bf{16303.8076}$ & $1542.7737$ & $\bf{16.2017}$ & $0.3320$ & $\bf{0.2067}$ & $0.0026$ \\
\hline
\end{tabular}}\label{table2}
\end{table}

\section{\label{sec:Discussion}Discussions}
We have proposed a novel data-driven ensemble method, based on deep neural networks, for modeling and prediction of chaotic dynamics consisting of extreme events as well as dynamics real-world scenarios. The proposed optimized ensemble deep learning framework is trained on time-series data and it requires no prior knowledge of the underlying governing equations. Using the trained model, long-term predictions are made by iteratively predicting one step forward. The proposed OEDL method is based on multiple deep models, the best convex optimization algorithm and an ensemble design strategy, useful for scalable forecasting. The proposal was first theoretically built as an optimization problem with possible solutions lying in a broader set of combinations of experts. The best convex combination of weights was chosen using an online solver which minimizes average squared loss and comes with finite time worst-case guarantees. This method is scalable to even larger systems, requires significantly less training data to obtain high-quality predictions in comparison with individual state-of-the-art deep learning forecasters, can effectively utilize an ensemble of base experts. Experimental results using both numerically simulated data sets and real-world data sets demonstrated the outstanding performance of the proposed OEDL method in comparison with other single and ensemble models. These results provide comprehensive evidence that the optimal ensemble deep learner is scalable for practical applications and leads to significant improvements over standard neural network methods. Since the discovery of the `Wisdom of Crowds' (more popularly `Vox Populi') over 100 years ago theories of collective intelligence have held that group accuracy requires either statistical independence or informational diversity among individual beliefs \cite{galton1907vox}. This idea received immense interest in time series forecasting when Bates and Granger \cite{bates1969combination} shows simple combinations of forecasting methods can generate better out-of-sample forecasts using empirical evidence. Of course, the current progress in machine learning and deep learning brought various highly capable forecasters to our door. This paper provides a mathematical optimization-based ensemble of deep learning methods for efficient and accurate forecasting of dynamics consisting of extreme events. Both theoretical and experimental results presented in this paper support our claims and improve the predictability of the dynamical systems considered in this study. 

A more principled way of ensemble learning by stacking is to perform it at the full-sequence level, rather than at the data level as attempted in this paper. The greater complexity of this pursuit leaves it to our future work. Also, with an increasingly large amount of training data and computing power, it is conceivable that the weight parameters and hyperparameters in all the constituent deep networks can be learned jointly using the unified backpropagation in an end-to-end fashion.

\begin{acknowledgments}
The authors would like to thank Chittaranjan Hens and Subrata Ghosh for constructive discussions and notable comments. 
\end{acknowledgments}

\section{APPENDIX}\label{sec:forecasting}

\section*{Neural network-based forecasting methods}
Deep learning systems have dramatically improved the accuracy of machine learning-based forecasting systems, and various deep architectures and learning methods have been developed with distinct strengths and weaknesses in recent years. Now, we discuss three state-of-the-art deep neural network frameworks for time series forecasting, which are used to develop the proposed OEDL framework. We have already mentioned that three expert forecasters are FFNN, RC, and LSTM. Above mentioned three types of neural network-based deep learning models are considered as basis time series models to construct the ensemble model, and their basic concepts and modeling process are briefly reviewed.

\subsection{Feed-forward neural network} 
\begin{figure}
	\includegraphics[scale=0.46]{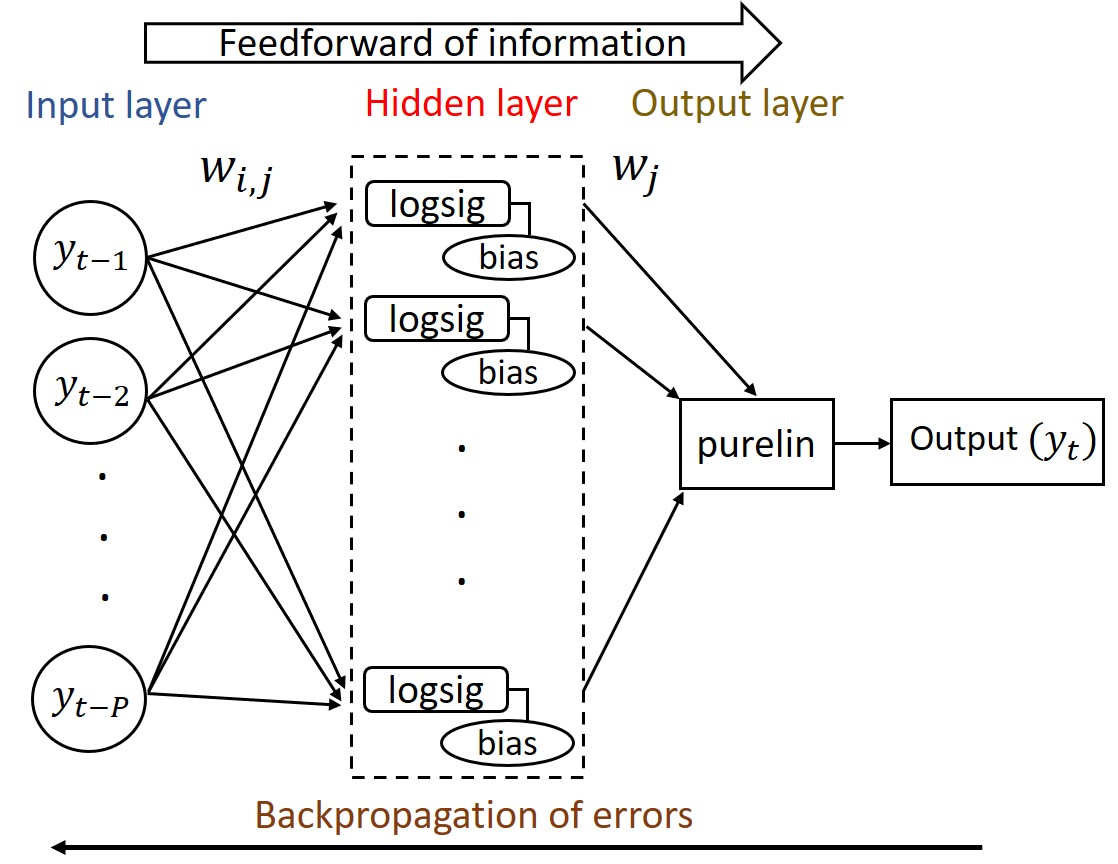}%
	\caption{A schematic diagram of a feed-forward neural network with one hidden layer consisting of $Q$-neurons and one output layer is drawn.  A set of $P$ inputs $\{y_{t-1},y_{t-2},\ldots,y_{t-P}\}$ is sent through input layer. The used activation functions are log-sigmoidal (denoted by 'logsig') for the hidden layer and linear for the output layer, respectively. The input weight parameters are $w_{i,j}$ whereas the output weight connections are $w_j$.}
	\label{figure_2} 
\end{figure}

Feed-forward neural networks (FFNN) are flexible computing frameworks
for modeling a broad range of nonlinear forecasting problems, inspired by the structure of human and animal brains. One significant advantage of the neural network models over other classes
of nonlinear models is that feed-forward neural network is universal
approximator that can approximate a large class of functions with
a high degree of accuracy \cite{hornik1989multilayer}. Their power comes from the parallel processing of the information from the data. No prior assumption on the data generating process is required in the model building process. Instead, the network model is largely determined by the characteristics of the data. Shallow feed-forward network with one hidden layer is the most widely used model for time series forecasting in various applied domains \cite{zhang1998forecasting}. A shallow FFNN model is characterized by a network of three layers of simple processing units connected by acyclic links (see Fig.\ \ref{figure_2}).

The relationship between the output $\left(y_{t}\right)$ and the inputs $\left(y_{t-1}, y_{t-2}, \ldots, y_{t-P}\right)$ is represented by the following mathematical equation,
\begin{equation}\label{ffnneqn}
y_{t}=w_{0}+\sum_{j=1}^{Q} w_{j}~g\left(w_{0 j}+\sum_{i=1}^{P} w_{i, j} y_{t-i}\right)+e_{t},
\end{equation}
where $w_{i, j} \; (i=0,1,2, \ldots, P, j=1,2, \ldots, Q)$ and $w_{j} \; (j=0,1,2, \ldots, Q)$ are
model parameters often called connection weights; $P$ is the number of input nodes; $Q$ is the number of hidden nodes, and $e_t$ be the error term. The logistic activation function is often used as the hidden layer transfer function, {\it i.e.}, $g(x)=\frac{1}{1+\exp (-x)}$.

Hence, the FFNN model as in Eqn. (\ref{ffnneqn}), in fact, performs a nonlinear functional mapping from the past observations to the future value $y_{t}$, i.e.,
$$y_{t}=f\left(y_{t-1}, \ldots, y_{t-P}, W\right)+e_{t},$$
where $W$ is a vector of all parameters and $f(\cdot)$ is a function determined by the network structure and connection weights. Thus, the neural network is equivalent to a nonlinear autoregressive model. Note that expression (\ref{ffnneqn}) implies one output node in the output layer, which is typically used for one-step-ahead forecasting. The neural network given by (\ref{ffnneqn}) can approximate arbitrary function when the number of hidden nodes $Q$ is sufficiently large \cite{zhang1998forecasting, hornik1989multilayer}. In practice, the FFNN structure with small number of hidden nodes often works well in out-of-sample forecasting. This may be due to the over-fitting effect that can be typically found in the neural network training process.

During the training phase of FFNN, it is a common technique to use a back-propagation learning algorithm \cite{rumelhart1985learning} for updating the weights and bias values by minimizing the error. Here error is the mean square of the difference between the predicted and actual outputs. We use one of the popularly used optimization algorithms, gradient descent with momentum \cite{qian1999momentum}, to minimize the error function. The momentum term may also be helpful to prevent the learning process from being trapped into a local minima and is usually chosen within the interval $[0, 1]$. Finally, the estimated model is evaluated using a separate hold-out sample that is not exposed to the training process.

\begin{figure}[t]
	\includegraphics[scale=0.31]{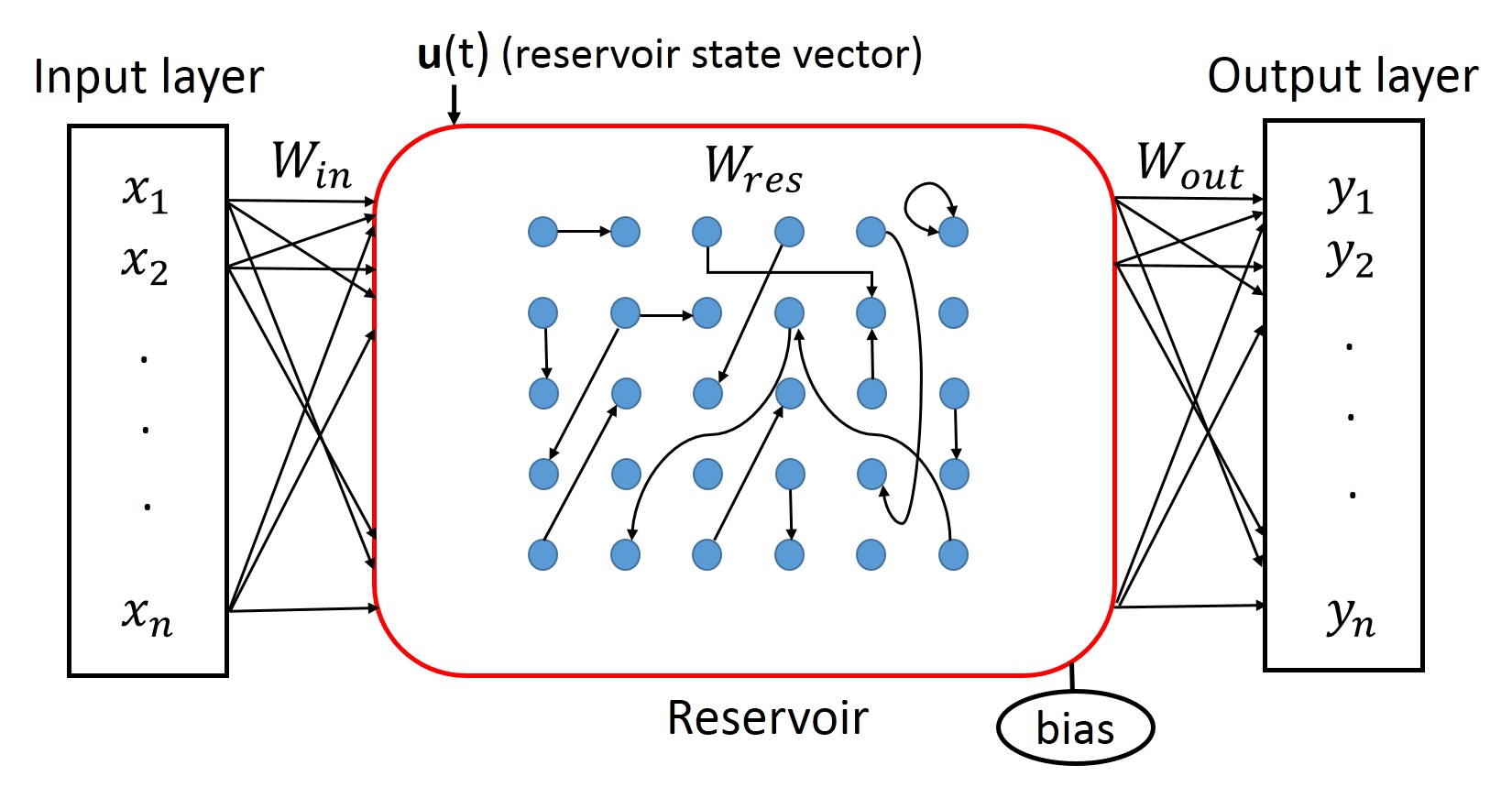}
	\caption{A schematic diagram of the reservoir computer consisting of input layer, reservoir, and output layer. For an input vector $\{x_1, x_2,\ldots, x_n\}$, a output vector $\{y_1, y_2,\ldots, y_n\}$ is calculated. Input weighted matrix and output weighted matrix are $[W_{in}]_{m \times n}$ and $[W_{out}]_{m \times n}$, respectively. $W_{res}$ is weighted random matrix of order $m \times m$ and ${\bf u(t)}$ is hidden state vector at $t$-th time.}
	\label{figure_3} 
\end{figure}

\begin{figure*}[t]
	\includegraphics[scale=0.55]{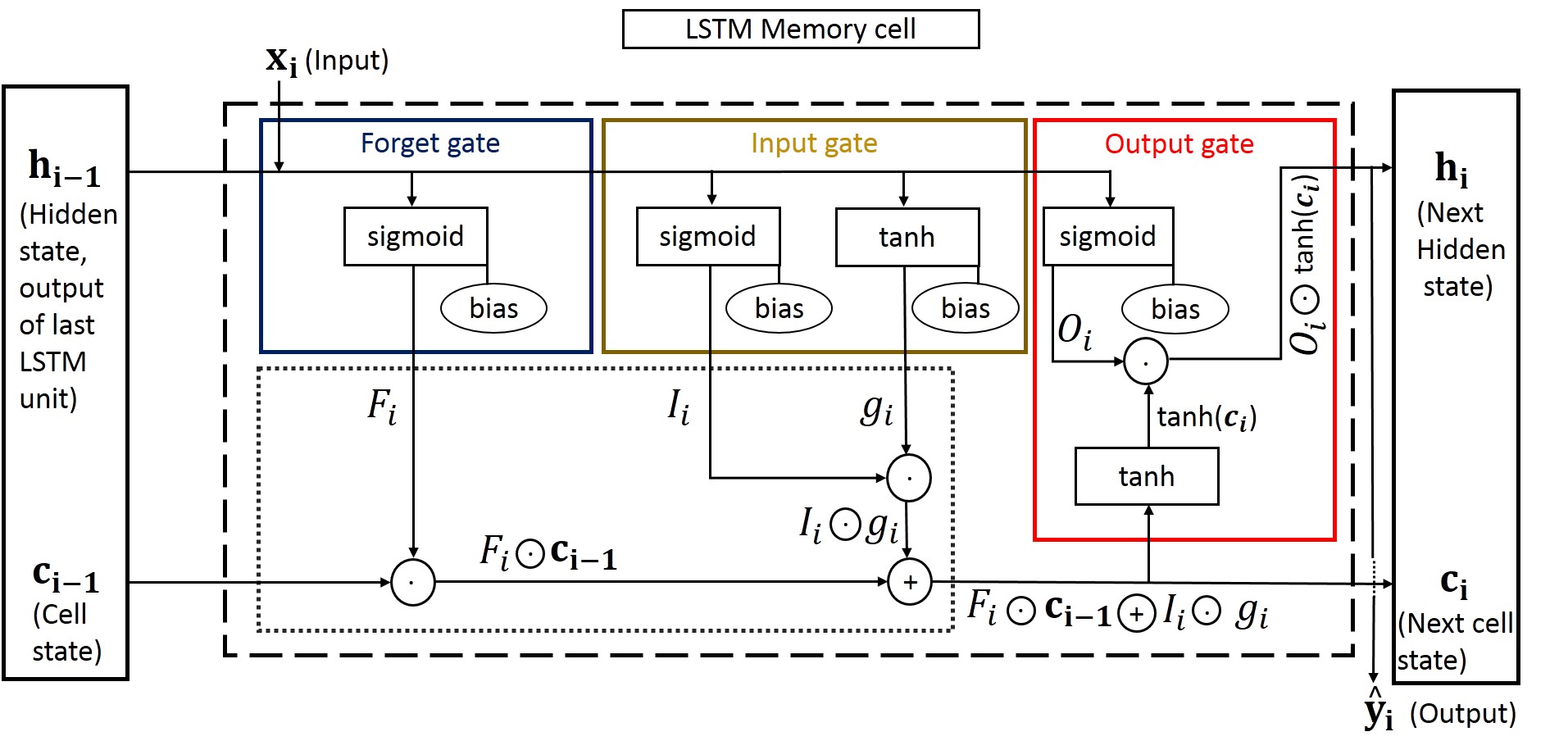}%
	\caption{A schematic diagram of the LSTM memory cell is portrayed. Updation of cell state and hidden state from $(i$-$1)$-th time to $i$-th time is shown through this diagram. The input vector  $\bf x_i$  is provided at $i$-th time. Three gates, i.e., forget gate, input gate, and output gate, are used for removing or adding information to the cell state. Hidden state and cell state at $i$-th time are denoted by $h_i$, and $c_i$. At the same time, $F_i$, $I_i$, $O_i$, and $g_i$ are forget gate's activation vector, input gate's activation vector, output gate's activation vector, and cell input activation vector, respectively. The hidden state vector $h_i$ contributes for getting output vector at $i$-th time.}
	\label{figure_4} 
\end{figure*}

\subsection{Reservoir computing} 
Training the FFNN model is a time-consuming process and also gives reduced accuracy in prediction. To overcome this, another powerful computational approach, namely Reservoir computing (RC), is developed based on recurrent neural networks. RC maps input signals into higher dimensional computational spaces through the dynamics of a fixed, non-linear system called a \textit{reservoir} \cite{schrauwen2007overview}. RC is a generalization of earlier neural network architectures such as recurrent neural networks, liquid-state machines, and echo-state networks \cite{jaeger2004harnessing}. Usage of RC is abundant at this time \cite{tanaka2019recent} and most recent applications of RC are evident in time series forecasting \cite{jaeger2002adaptive,wyffels2010comparative}, robot motor control \cite{salmen2005echo}, speech recognition \cite{jalalvand2011connected} among many others. The basic architecture of the reservoir computer for computation is delineated below. 

The reservoir computer consists of an input layer, a reservoir, and an output layer. A schematic diagram of it is portrayed in Fig.\ \ref{figure_3}. A input vector ${\bf x}=\{x_1,x_2,\ldots,x_n\}^T$ of $n$-dimension is sent for training from input layer into reservoir using the linear input weighted matrix $W_{in}$ of order $m\times n$, whose elements lie within $[-1,1]$. An $m$-dimensional reservoir state vector ${\bf u}(t)$ (or hidden state vector) is defined for reservoir neuron activation and ${\bf u}(t)$ follows the following deterministic model, 
\[
{\bf u}(t+\Delta t)=(1-\beta){\bf u}(t)+\beta \tanh(W_{res}{\bf u}(t)+W_{in}{\bf x}(t)+W_{b}),
\] 
where $\Delta t$ is the step of time discretization. For this study, we use $\Delta t=1$. Here, $W_{res}$ denotes the recurrent weighted adjacency matrix of order $m\times m$ of the reservoir. This matrix is a sparse random Erd\"{o}s-R\'{e}nyi matrix whose non-zero entities are picked up from $[-1, 1]$ uniformly. All the elements of $W_{res}$ are rescaled uniformly in such a way that the largest value of the magnitudes of its eigenvalues becomes a pre-defined positive real number, called the \textit{`spectral radius'} of $W_{res}$. The leakage rate is denoted by $\beta$ that lies in $(0,1]$. We set the spectral radius and the leaking rate as $0.001$, and $0.5$, respectively for experimental usage. The bias vector of order $m\times 1$ is denoted by $W_b$ whose elements are $1$. Again a linear output weighted matrix $W_{out}$ of order $m\times n$ is constructed for generating the output vector ${\bf \hat y}=\{y_1,y_2,\ldots,y_n\}^T$ at $t$-th time in the following way,
$${\bf \hat y}(t)=W^T_{out}{\bf u}(t).$$
Here, elements of $W_{out}$ are adjusted in the training phase from the reservoir using linear regression which is ridge regression \cite{lukovsevivcius2009reservoir}. This ridge regression minimizes the mean square error of the network output data and the training target data and the output weights are computed by the following way,
$$W^T_{out}={\bf Y}_{target}{\bf X}^T({\bf X}{\bf X}^T+\beta^2 {\bf I})^{-1}.$$
Here, ${\bf Y}_{target}$ and ${\bf X}$ consist of target output data and reservoir state collected from each training time iteration. Let $p$ be the training time steps. Then orders of ${\bf Y}_{target}$ and ${\bf X}$ are $n \times p$ and $m \times p$, respectively. Since the concept of reservoir computing stems from the use of recursive connections within neural networks to create a complex dynamical system; thus it is highly useful in the forecasting of dynamics \cite{lu2017reservoir, pathak2018hybrid, chen2020autoreservoir}.

\subsection{Long short-term memory}
To overcome the problems of speed and stability in recurrent neural network, long short-term
memory (LSTM) was proposed by Hochreiter and Schmidhuber \cite{hochreiter1997long}. The LSTM is similar to the recurrent neural network, but in this model, a new concept is introduced with one cell or interaction per module. As presented in Fig.\ \ref{figure_4}, LSTM is a chain-like structure capable of remembering information and long-term training with four network layers. Besides, LSTM is designed in such a way that the vanishing gradient problem is got over. For this, usage of LSTM became a popular choice in various applied fields \cite{yu2019review,shastri2020time}. 

The LSTM network consists of memory blocks, like cells. As usual, a hidden state is introduced in LSTM. Besides, the memory of LSTM is carried by a cell state and the cell state is updated by removing or adding information when it passes through three different gates, namely forget gate, input gate, and output gate. With the implementation of these gates, long-term dependency problems can be avoided while memorizing the LSTM. Now, we focus on how the cell and hidden states are modified corresponding to the input vector (say, $\bf x_{i}\in \mathbb{R}^n$) at $i$-th time step.

{\it Forget gate:} The forget gate concludes about the information that is removed in the cell state $\bf c_i$ at $i$-th time step. Let $\bf h_{i-1}\in \mathbb{R}^m$ be the hidden state at previous $(i$-$1)$-th step. Due to implementation of sigmoid activation function ($\sigma$) over the combination of $\bf x_i$ and $\bf h_{i-1}$ with their corresponding weighted matrices $W_{x1}$, and $W_{h1}$, respectively and bias vector $b_1\in \mathbb{R}^m$, a forget gate's activation vector $F_i\in \mathbb{R}^m$ is generated in the following way: 
\[
F_i=\sigma(W_{x1} {\bf x_i}+W_{h1} {\bf h_{i-1}}+b_1).
\]
Here, the orders of the matrices $W_{x1}$, and $W_{h1}$ are $m\times n$ and $m\times m$, respectively. It gives the result lying within $[0, 1]$, where $0$ indicates forgetting the information completely, whereas $1$ signifies to keep the information as it is.

{\it Input gate}: Input gate decides how much the new information is added to the cell state.
Input gate's activation vector $I_i\in \mathbb{R}^m$ is produced by using a sigmoid activation function as follows:
$$
I_i=\sigma(W_{x2} {\bf x_i}+W_{h2} {\bf h_{i-1}}+b_2).$$ 
Besides, a vector of new candidate values $g_i\in \mathbb{R}^m$ is created through $\tanh$ activation function as follows:
$$
g_i=tanh(W_{x3} {\bf x_i}+W_{h3} {\bf h_{i-1}}+b_3).$$
Here the orders of $W_{x2}$ and $W_{x3}$ are $m\times n$, the orders of $W_{h2}$ and $W_{h3}$ are $m\times m$, and $b_2, b_3\in \mathbb{R}^m$ are bias vector. $I_i\odot g_i$ contributes for updating cell state where the operator $\odot$ represents the point-wise multiplication between two vectors.

Finally, the linear combination of the input gate and forget gate is employed for updating the previous cell state into the current cell state by the following equation:
$${\bf c_i}=F_i\odot {\bf c_{i-1}}\oplus I_i\odot g_i.$$

{\it Output gate}: Finally, the hidden state is modified in the output gate. Output gate's activation vector is determined in the following way:
\[
O_i=\sigma(W_{x4} {\bf x_i}+W_{h4} {\bf h_{i-1}}+b_4),
\]
where $W_{x4}\in \mathbb{R}^{m\times n}$, $W_{h4}\in \mathbb{R}^{m\times m}$, and $b_{4}\in \mathbb{R}^m$ are weighted matrices and bias vector, respectively. The current hidden state is updated by operating $\tanh$ activation function over current cell state as follows:
\[
{\bf h_i}=O_i\odot tanh({\bf c_i}),
\]
where the initial values are setted as $\bf c_0 = 0$, and $\bf h_0=0$.
So, the output vector is calculated from the hidden state at the time step $i$ as follows
\[
{\bf \hat y_i}=\sigma(W {\bf h_{i}}+b_5),
\]
where $\sigma$ is activation function, $W\in \mathbb{R}^{n\times m}$ is weighted matrix and $b_{5}\in \mathbb{R}^n$ is bias vector. As a result of this state transition within LSTM, the trained LSTM network with a set of fixed weight vectors can still show different predictive behaviour for different time series.


\bibliography{ml_forecast}

\end{document}